\documentclass[11pt,a4paper]{article}

\usepackage[margin=1in]{geometry}
\usepackage{hyperref}
\usepackage{booktabs}
\usepackage{graphicx}
\usepackage{array}
\usepackage{amsmath}
\usepackage{amssymb}
\usepackage{url}
\usepackage{float}
\usepackage{algorithm}
\usepackage{algpseudocode}
\usepackage[protrusion=true,expansion=false]{microtype}

\hypersetup{colorlinks=true,linkcolor=black,urlcolor=blue,citecolor=blue}

\PassOptionsToPackage{hyphens}{url}
\makeatletter
\g@addto@macro\UrlBreaks{\UrlOrds}
\g@addto@macro\UrlBreaks{\do\_\do\/\do\-\do\.\do\:\do\(\do\)\do\[\do\]\do\,}
\makeatother

\newcommand{\plusminus}{$\pm$}
\title{ConvMemory v3: A Validity Context Layer for Conversational
       Memory \\ via Target-Conditioned Relation Verification}

\author{Taiheng Pan \\
        School of Computing and Information Systems \\
        University of Melbourne \\
        \texttt{github.com/pth2002}}

\date{June 2026}

\begin{document}

\maketitle

\begin{abstract}
Conversational memory retrieval optimizes relevance, yet a retrieved memory can
be relevant and simultaneously outdated: a later turn updates, corrects, or
supersedes it. ConvMemory v3 adds a \emph{validity context layer} that detects
and surfaces this update evidence through target-conditioned relation
verification, sitting after the v1/v2 retrieval path. The core mechanism is a
dual-evidence gate that conditions a relation judgment on the specific target
proposition, scoring a (target, source) pair through the product of a MiniLM slot
head and a DeBERTa-v3 slot head and gating it by conservative event/operation
evidence. On a synthetic multi-hop validity benchmark the gate reaches
$90.12\% \pm 1.73$ accuracy; through a real-data feedback loop that mines failure
patterns but trains on synthetic pairs only, the verifier
transfers to Memora role binding with zero target-side labels, reaching
$98.8\% \pm 0.9$ group-all-correct. The deployed layer preserves retrieval by
default: a \texttt{context} mode attaches structured validity metadata while
keeping the candidate set and rank order fixed, and a query-conditioned
\texttt{demote} mode is an explicit opt-in for dense current-state workloads, where
it raises current-active H@1 from a never-demote baseline of $45.1\%$ to
$95.7\% \pm 1.2$ while protecting non-superseded memories at $99.4\%$ recall. Six
machine-verifiable safety contracts pin the layer's behavior. Multi-hop graph
propagation is validated as a mechanism; fully automatic construction of strict
prerequisite edges is characterized as a boundary, since strict necessity requires
counterfactual world knowledge. This report extends ConvMemory v1
\cite{convmemory_v1} and v2 \cite{convmemory_v2}.
\end{abstract}

\newpage
\setcounter{tocdepth}{2}
\tableofcontents
\newpage

\section{Introduction}
\label{sec:intro}

A conversational agent accumulates memories across many sessions, and a query
against that store asks not only ``which memory is relevant'' but increasingly
``which memory is still true.'' A user who changes jobs, moves cities, or revises
a preference leaves a trail in which an earlier memory remains topically relevant
while a later memory supersedes it. ConvMemory v1 \cite{convmemory_v1} and v2
\cite{convmemory_v2} both optimize relevance: v1 organizes a high-recall top-500
pool with a lightweight learned reranker, and v2 adds a recall-preserving top-10
evidence reranker that improves ordering inside the protected set. Neither layer
reasons about whether a relevant memory has been overturned.

ConvMemory v3 adds that reasoning as a \emph{validity context layer}. The layer
runs after the v1/v2 retrieval path and answers a different question for each
retrieved memory: relative to the specific target the query is about, does a
later memory update, correct, or supersede this one? The answer is attached as
structured validity metadata, and the agent decides how to use it. By default the
\texttt{context} mode preserves the candidate set and rank order while extending
the result schema with validity metadata; the v1/v2 retrieval output is unchanged,
and the \texttt{off}/\texttt{None} mode is byte-identical to a build without v3.

The technical core is a \emph{target-conditioned relation verifier}. A relation
judgment that ignores the target collapses on the cases that matter: two memories
about the same person can stand in an update relation for one attribute and an
unrelated relation for another, and the relation is only well-defined once the
target proposition is fixed. v3 conditions the judgment on the target through a
dual-evidence gate that multiplies a MiniLM slot score by a DeBERTa-v3 slot
score, gates the product by conservative event/operation evidence, and propagates
the result over sources by noisy-or. A stratified-supervision feedback loop learns
from real role-binding hard cases and transfers back with zero target-side labels.

We make three contributions.

\begin{enumerate}
    \item \textbf{A target-conditioned dual-evidence gate} (the mechanism): a
    slot-product verifier with a conservative event/operation gate that reaches
    $90.12\% \pm 1.73$ on a synthetic multi-hop validity benchmark and, through a
    real-data feedback loop, transfers to Memora role binding at $98.8\% \pm 0.9$
    group-all-correct with zero target-side labels. This is well above a
    target-position rule ($78.6\%$), zero-shot NLI ($64.2\%$), and a relevance
    cross-encoder ($17.9\%$).
    \item \textbf{A validity context layer} (the system): an opt-in stage exposing
    four accepted values through a single \texttt{validity\_mode} argument
    (\texttt{None}, \texttt{off}, \texttt{context}, \texttt{demote}), corresponding
    to three behavioral modes. The
    \texttt{context} default preserves the candidate set and rank order while
    attaching validity evidence; the \texttt{demote} mode is an explicit opt-in
    that reorders for dense current-state workloads, raising current-active H@1
    from $45.1\%$ to $95.7\% \pm 1.2$ on Memora dense retrieval while protecting
    non-superseded memories at $99.4\%$ recall. Six machine-verifiable safety
    contracts pin the layer's behavior, backed by 41 passing tests.
    \item \textbf{Two boundary findings} (the scope): multi-hop graph propagation
    is validated as a mechanism with given structure; target-conditioned relation
    labels for free-form multi-hop conversation are, to our knowledge, scarce, and fully
    automatic construction of strict prerequisite edges requires counterfactual
    necessity judgment that a discriminative model does not provide reliably. Both
    are characterized precisely in \S\ref{sec:findings}.
\end{enumerate}

v3 builds directly on the ConvMemory cascade: v1 supplies the candidate pool, v2
supplies the protected ordering, and v3 supplies validity context. Two models
carry distinct roles, and the report keeps them separate throughout. The
dual-evidence verifier establishes that target-conditioned relation verification
works, and supplies the synthetic and role-binding evidence for the mechanism. The
deployed demotion calibrator is the query-conditioned model the released package
runs, and it produces the dense-retrieval current-active H@1 of $95.7\%$. The
verifier, its training, the propagation mechanism, and the deployed calibrator are
described in turn, each with the experiments that establish it.

\section{Relationship to the v1 and v2 Reports}
\label{sec:rel}

ConvMemory v1 \cite{convmemory_v1} introduced a lightweight learned memory
reranker, a negative attribution result on a learned temporal window, and a
research-preview conflict editor. ConvMemory v2 \cite{convmemory_v2} introduced a
recall-preserving top-10 evidence reranker that reorders only v1's protected
top-10 and improves MRR and H@1 while keeping Recall@10 fixed by construction. v3
composes with both rather than replacing either.

\paragraph{v1 (the conflict-editor research preview).} v1 shipped a
research-preview conflict editor and listed conflict modeling as open. v3 is the
matured successor to that direction: instead of editing conflicting memories in
place, v3 verifies the relation between a target memory and a later source and
surfaces the result as validity context. The v3 verifier is target-conditioned
and supervised, where the v1 preview was neither.

\paragraph{v2 (the recall-preserving cascade).} v2 reorders v1's protected
top-10 and leaves recall fixed. v3 attaches after that cascade. In the
\texttt{context} default, v3 preserves whatever ordering v1/v2 produced and only
adds validity metadata, so a v1+v2+v3 deployment retains v2's exact retrieval
behavior. The \texttt{demote} opt-in is the single mode in which v3 reorders, and
it is scoped to dense current-state workloads (\S\ref{sec:dense}).

\paragraph{The temporal-window negative result.} v1 reported that a learned
temporal window was statistically significant on aggregate but not temporally
specific, and v2 made no temporal-mechanism assertion. v3's mechanism is
target-conditioned relation verification on memory text, distinct from a temporal
window: the validity signal comes from what two memories assert about a shared
target, not from their position in time. The v1 negative result stands unchanged.

\section{Related Work}
\label{sec:related}

v3 sits at the intersection of three lines of work, and its contribution is
clearest when contrasted with each.

\paragraph{Agent memory systems.} A growing body of systems give conversational
agents persistent memory. Systems such as MemGPT \cite{memgpt2023}, Generative
Agents \cite{genagents2023}, MemoryBank \cite{memorybank2024}, and Mem0
\cite{mem02025} write summaries or facts to a store, index them, and retrieve them
by relevance when a later query needs history. Their objective is coverage and
recall, making sure the right memory can be found again. v3 is orthogonal and
complementary: it does not change what is stored or how it is retrieved, but adds a
layer that judges, for a retrieved memory, whether a later memory has superseded it
with respect to the target the query asks about. Storage and retrieval answer
``can this memory be found''; the validity layer answers ``is this memory still
true.'' Benchmarks for long-term conversational memory, such as LoCoMo
\cite{locomo2024} and LongMemEval \cite{longmemeval2024}, mostly frame the task as
fact retrieval from past sessions; Memora \cite{memora2026} is closest to v3 in
foregrounding memory mutation and obsolescence, which is the regime the validity
layer targets.

\paragraph{Knowledge editing.} Knowledge-editing methods such as ROME
\cite{rome2022} and MEMIT \cite{memit2022} update what a model knows
by editing its weights, so that the model itself stops asserting an outdated fact.
v3 leaves model weights untouched and operates on the retrieval context instead: it
annotates or reorders the retrieved memories that a downstream model will read.
Editing changes the model; the validity layer changes the evidence the model is
given, which keeps the original memories auditable and the layer removable.

\paragraph{Natural language inference and fact verification.} NLI and
fact-verification models, trained on datasets such as FEVER \cite{fever2018}, judge
whether one sentence entails, contradicts, or
supports another. v3 uses an NLI backbone for one of its slot heads, but the task
is different: fact verification asks whether two sentences are consistent in
isolation, whereas the validity layer asks whether a source overturns a target
\emph{with respect to a specific proposition}. A sentence-level contradiction
signal that ignores the target conflates an employer update with an unrelated
hometown statement; the target-conditioned slot product is precisely what a
query-blind NLI judgment lacks (\S\ref{sec:dense}).

The components v3 builds on are concrete. The verifier uses two slot heads: a
MiniLM cross-encoder \cite{minilm2020}\footnote{\texttt{cross-encoder/ms-marco-MiniLM-L6-v2}.}
of the family v1 used as a distillation
teacher and v2 used as its evidence scorer, fine-tuned on MS MARCO passage ranking
\cite{msmarco2016, sbert2019}, and a DeBERTa-v3 backbone \cite{debertav3}
initialized from an NLI checkpoint\footnote{\texttt{cross-encoder/nli-deberta-v3-base},
based on \texttt{microsoft/deberta-v3-base} and trained on SNLI and MultiNLI.} and
fine-tuned with a binary classification
head. The retrieval path beneath v3 is the ConvMemory v1 reranker over a dense
MPNet \cite{mpnet2020} top-500 pool, optionally followed by the v2 evidence
reranker. Evaluation uses three settings: a synthetic multi-hop validity benchmark
constructed for this work, the Memora role-binding transfer setting, and Memora
dense current-state retrieval. The recall-preserving cascade and the protected
top-10 are defined in the v1 and v2 reports \cite{convmemory_v1, convmemory_v2}; v3
treats them as the fixed substrate it annotates.

\section{The Validity Problem}
\label{sec:problem}

\paragraph{Relevance is not validity.} Given a query about a target entity and an
accumulated memory store, a retriever ranks memories by relevance. A memory can
rank high on relevance and still be outdated: the user has since updated the fact.
The validity question is orthogonal: among the relevant memories, which ones a
later source memory has superseded with respect to the target the query asks
about.

\paragraph{The relation is target-conditioned.} The relation between two memories
is only well-defined once the target proposition is fixed. Two statements about
the same person can stand in an update relation on one attribute (employer) and
an unrelated relation on another (hometown). A relation judgment that ignores the
target conflates these. v3 therefore conditions every relation judgment on the
specific target, which is the property that separates it from a query-blind
relation classifier (\S\ref{sec:dense}).

\paragraph{Two regimes.} Validity reasoning takes two forms in practice. In
\emph{role binding}, a single later memory directly rebinds a target's attribute,
and the verifier must decide whether a candidate is the superseded binding. In
\emph{multi-hop propagation}, an update at one node invalidates a chain of
downstream memories that depended on it, and validity must propagate along the
dependency structure. v3 addresses role binding as its load-bearing, real-data
validated capability and validates multi-hop propagation as a mechanism with
given structure, with a documented data boundary (\S\ref{sec:findings}).

\section{The Target-Conditioned Dual-Evidence Gate}
\label{sec:verifier}

\subsection{Slot heads}
\label{sec:slots}

The verifier scores whether a source memory \emph{directly and specifically}
overturns a target memory. The central quantity is the \emph{slot score}: a
target-conditioned proposition score that measures whether the source acts on the
specific target proposition (role binding), rather than merely sharing a topic.
The slot score is produced by two heads that share an input form but use different
backbones.

Both heads receive the source side and the target side rendered as a text pair.
The source side is
\begin{quote}\small
\texttt{SOURCE\_MEMORY: \{source\_text\}} \\
\texttt{QUESTION: Does SOURCE directly and specifically overturn TARGET?}
\end{quote}
paired with the target side
\begin{quote}\small
\texttt{TARGET\_MEMORY: \{target\_text\}} \\
\texttt{DOMAIN: \{domain\}}
\end{quote}
The \textbf{MiniLM head} is a binary cross-encoder over this pair, producing
$m_{\mathrm{mini}} = P(\text{direct\_overturn})$. The \textbf{DeBERTa-v3 head}
takes the same input form, but its backbone is initialized from an NLI checkpoint
and the three-way entailment head is replaced by a binary classification head, so
it produces a comparable $m_{\mathrm{nli}} = P(\text{direct\_overturn})$ rather
than raw entail/neutral/contradict logits. The slot score is the product of the
two heads:
\begin{equation}
\mathrm{slot} \;=\; m_{\mathrm{mini}} \cdot m_{\mathrm{nli}}.
\label{eq:slot}
\end{equation}
The product form is conservative: both heads must agree that the source
specifically overturns the target proposition for the slot score to be high.

\subsection{Conservative event/operation gate}
\label{sec:gate}

\begin{figure}[htbp]
\centering
\includegraphics[width=0.92\textwidth]{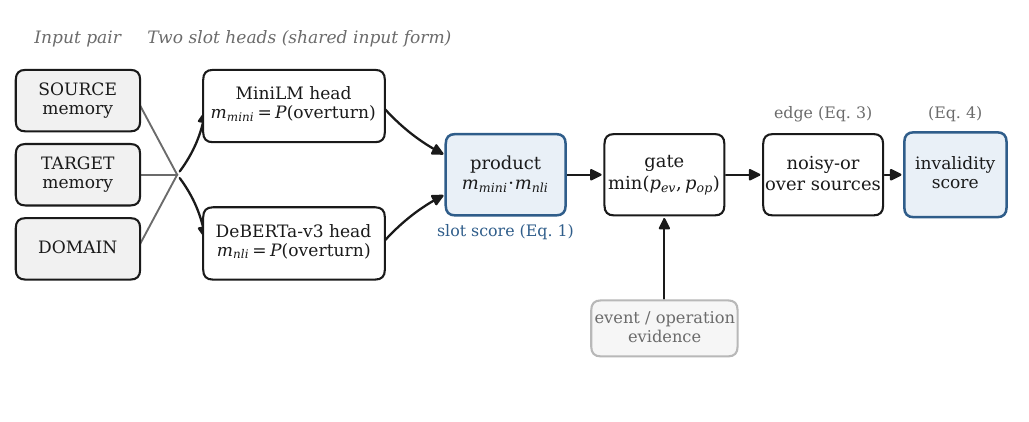}
\caption{The target-conditioned dual-evidence gate. A (target, source) pair, with
domain, is scored by two slot heads (MiniLM and DeBERTa-v3) sharing
the same input form; their product is the target-conditioned slot score
(Equation~\eqref{eq:slot}). The slot score is gated by the conservative
$\min(p_{ev}, p_{op})$ factor (Equation~\eqref{eq:edge-new}) to form an edge
score, and edge scores are combined over sources and over the sample by noisy-or
(Equation~\eqref{eq:noisyor}) into an invalidity score.}
\label{fig:gate}
\end{figure}

The slot score answers ``does the source act on this target proposition.'' Two
further factors answer ``is the source an update event at all'' and ``is the
operation an overturn.'' Let $p_{ev} = P(\text{source is an update/conflict
event})$ and $p_{op} = P(\text{operation type is overturn})$. These two factors
are produced by an event head and an operation head from the earlier three-head
attachment model, separate from the two slot heads of \S\ref{sec:slots}. The
original formulation multiplied three independent factors,
\begin{equation}
\mathrm{edge} \;=\; p_{ev} \cdot \mathrm{slot} \cdot p_{op},
\label{eq:edge-old}
\end{equation}
with the slot factor itself the product of Equation~\eqref{eq:slot}. A merge
analysis (\S\ref{sec:merge}) showed that the two noisy event/operation heads need
not be multiplied independently: collapsing them into a
single conservative gate is statistically equivalent, and the frozen dual-gate form is
\begin{equation}
\mathrm{edge} \;=\; \mathrm{slot} \cdot \min(p_{ev},\, p_{op}).
\label{eq:edge-new}
\end{equation}
The $\min$ gate is conservative in the same spirit as the slot product: the edge
fires only when the weakest of the two evidence factors supports it.
Figure~\ref{fig:gate} shows the full data flow.

\subsection{Aggregation by noisy-or}
\label{sec:noisyor}

A target may be overturned by any of several sources. Edge scores from the same
source are combined, and source scores are combined over the sample, by noisy-or:
\begin{equation}
\mathrm{noisy\_or}(\{x_i\}) \;=\; 1 - \prod_i (1 - x_i),
\qquad
\begin{aligned}
\mathrm{source\_score} &= \mathrm{noisy\_or}(\text{edges of one source}), \\
\mathrm{invalid\_score} &= \mathrm{noisy\_or}(\text{source scores}).
\end{aligned}
\label{eq:noisyor}
\end{equation}
The phrase ``full product $+$ noisy-or'' denotes this composition precisely: the
MiniLM and DeBERTa heads are combined by \emph{product}
(Equation~\eqref{eq:slot}), and the resulting edges are combined across sources by
\emph{noisy-or} (Equation~\eqref{eq:noisyor}). The two heads are never combined by
noisy-or with each other. A per-seed, per-fold decision threshold is selected on a
validation split, optimizing overall accuracy jointly with invalidated-case and
valid-case accuracy. Algorithm~\ref{alg:gate} states the full scoring path.

\begin{algorithm}[htbp]
\caption{Dual-evidence gate: target invalidity score}
\label{alg:gate}
\begin{algorithmic}[1]
\Require target memory $t$; candidate sources $\mathcal{S}$; domain $d$
\Ensure invalidity score for $t$
\State $\textit{source\_scores} \gets [\,]$
\For{each source $s \in \mathcal{S}$}
    \State $x \gets$ render pair $(s, t, d)$ in the slot input form
    \State $m_{\mathrm{mini}} \gets$ MiniLM head$(x)$
    \Comment{$P(\text{direct\_overturn})$}
    \State $m_{\mathrm{nli}} \gets$ DeBERTa-v3 head$(x)$
    \Comment{NLI backbone, binary head}
    \State $\mathrm{slot} \gets m_{\mathrm{mini}} \cdot m_{\mathrm{nli}}$
    \State $p_{ev} \gets$ event head$(x)$
    \Comment{$P(\text{update/conflict event})$}
    \State $p_{op} \gets$ operation head$(x)$
    \Comment{$P(\text{operation is overturn})$}
    \State $\mathrm{edge} \gets \mathrm{slot} \cdot \min(p_{ev}, p_{op})$
    \State append $\mathrm{noisy\_or}(\{\mathrm{edge}\})$ to $\textit{source\_scores}$
\EndFor
\State \Return $\mathrm{noisy\_or}(\textit{source\_scores})$
\end{algorithmic}
\end{algorithm}

\subsection{A worked example}
\label{sec:worked}

A concrete case makes the gate's behavior explicit. The scores below come from the
graph-line edge-score probe, the four-head diagnostic scorer of
Equations~\eqref{eq:slot}--\eqref{eq:edge-old}, which exposes the component heads
that the released single calibrator folds together (\S\ref{sec:calibrator}). Take
the source evidence
\emph{``Policy Z12 eligibility criteria revised; claims must now include additional
documentation''} and the target memory \emph{``Claim \#5241 meets eligibility
criteria under Policy Z12.''} The source is an update event acting by overturn, so
the event and operation factors are high ($p_{ev} = 0.9968$, $p_{op} = 0.9933$).
Both slot heads agree that the source specifically overturns the target
proposition ($m_{\mathrm{mini}} = 0.99995$, $m_{\mathrm{nli}} = 0.99888$), so the
slot product is near unity ($\mathrm{slot} \approx 0.9988$). The two edge forms
agree on this case: the four-factor diagnostic
$p_{ev}\cdot p_{op}\cdot m_{\mathrm{mini}}\cdot m_{\mathrm{nli}} \approx 0.9890$,
and the frozen merged gate
$\mathrm{slot}\cdot\min(p_{ev}, p_{op}) \approx 0.9921$, both place the edge near
unity, matching the no-measurable-loss merge of \S\ref{sec:merge}.
The target is surfaced as \texttt{possibly\_outdated} in \texttt{context} mode and
is demotable under the explicit \texttt{demote} opt-in. The discriminating factor
is the slot product: a source that were topically related but did not act on this
specific eligibility proposition would score low on the slot heads, driving the
product toward zero regardless of the event factor. This is the scope-mismatch hard
negative that stratified supervision targets (\S\ref{sec:strat}).

\subsection{Architecture freeze: head merge and the load-bearing spine}
\label{sec:merge}

Two architectural questions were settled by paired ablation with 95\% confidence
intervals on the synthetic benchmark.

\paragraph{The event and operation heads merge with no measurable loss.} Replacing the
independent product $p_{ev}\cdot p_{op}$ with the conservative gate
$\min(p_{ev},p_{op})$ of Equation~\eqref{eq:edge-new} changes accuracy by
$+0.00\%$ with 95\% CI $[-0.88\%, +0.84\%]$. The merged gate is the frozen dual-gate
architecture: simpler and statistically indistinguishable from the two-head form.

\paragraph{The slot product is the spine.} Scoring with the slot-only product and
noisy-or gives $-1.00\%$ relative to the full configuration (95\% CI
$[-2.19\%, +0.20\%]$), within noise: the slot product carries essentially all
of the signal. Removing the slot mechanism entirely (event/operation heads without
the slot interaction) costs about 14 percentage points. The target-conditioned
slot product is therefore the load-bearing spine, and the event/operation heads
are mergeable, conservative overhead. Table~\ref{tab:arch} collects the ablation.

\begin{table}[htbp]
\centering
\caption{Architecture ablation on the synthetic multi-hop benchmark (paired, 95\%
CI). The merged event/operation gate matches the full two-head model; the
slot-only product matches the full configuration within noise; removing the slot
interaction costs about 14 points. The merged-gate, slot-spine architecture is the frozen research-verifier
configuration.}
\label{tab:arch}
\renewcommand{\arraystretch}{1.2}
\begin{tabular}{>{\raggedright\arraybackslash}p{0.55\textwidth} r}
\toprule
Configuration (vs full) & Paired delta (95\% CI) \\
\midrule
Merged $\min(p_{ev},p_{op})$ gate vs full two-head & $+0.00\%$ \,$[-0.88\%, +0.84\%]$ \\
Slot-only product $+$ noisy-or vs full              & $-1.00\%$ \,$[-2.19\%, +0.20\%]$ \\
Event/operation heads without the slot interaction & about $-14$ pp \\
\bottomrule
\end{tabular}
\end{table}

\section{Training the Verifier}
\label{sec:training}

\subsection{Objective}
\label{sec:objective}

Each slot head is trained as a binary classifier on labeled (source, target,
domain) pairs. A two-logit head uses weighted cross-entropy; a single-logit head
uses binary cross-entropy with logits:
\begin{equation}
\mathcal{L}_{\mathrm{slot}} \;=\; \mathrm{CE}_{w}\big(\hat{y},\, y\big),
\qquad y \in \{\text{overturn},\ \text{non-overturn}\}.
\label{eq:loss-slot}
\end{equation}
The class weighting addresses the imbalance between the relatively rare direct
overturns and the abundant non-overturns.

\subsection{Stratified supervision}
\label{sec:strat}

The discriminating signal lives in the hard cases, so the training distribution is
stratified rather than sampled uniformly:
\begin{itemize}
    \item \textbf{Clean positive}: a GPT-confirmed direct text overturn.
    \item \textbf{Hard negative}: scope mismatch or temporal non-conflict, namely a
    source that looks superficially like an overturn but does not act on the
    target proposition.
    \item \textbf{Easy negative}: an ordinary unrelated pair.
\end{itemize}
Clean positives are confirmed by a GPT labeler (\texttt{gpt-4.1}) that judges whether the source
directly overturns the target. These labels supervise training only; no GPT label
enters the evaluation sets, whose targets come from the synthetic-benchmark
construction and the Memora ground truth.
Hard negatives are oversampled by a focus multiplier (default
$\textit{hard\_focus\_multiplier}=4$), and low-overlap or failure-subtype
positives are oversampled (default $\textit{pos\_multiplier}=3$). Stratifying the
signal so that rule-easy and rule-hard cases contribute distinct gradients, rather
than being averaged, produced the first large jump to $84.5\% \pm 2.8$
(\S\ref{sec:progression}).

\subsection{The real-data feedback loop}
\label{sec:feedback}

The transfer capability comes from a feedback loop that analyzes real failure
patterns, generates synthetic supervision matching them, and transfers back with
zero target-side labels. Algorithm~\ref{alg:loop} states the loop.

\begin{algorithm}[htbp]
\caption{Real-data feedback loop (zero target-side labels at transfer)}
\label{alg:loop}
\begin{algorithmic}[1]
\State Analyze Memora role-binding failures of a target-position rule and
       zero-shot NLI to identify recurring hard-case patterns
\State Generate synthetic paired supervision (clean positive vs scope/temporal
       hard negative) matching those patterns, without Memora labels
\State Refine the slot heads on the synthetic pairs under the stratified objective
       (\S\ref{sec:strat})
\State Evaluate transfer on Memora with no target-side training labels
\State \Return refined verifier
\end{algorithmic}
\end{algorithm}

This loop is the strongest validated capability in v3: a verifier trained without
Memora target labels transfers to the full Memora role-binding set at
$98.8\% \pm 0.9$ group-all-correct (\S\ref{sec:rolebinding}).

\paragraph{What zero target-side labels means.} The transfer is label-free in a
precise sense: training uses synthetic pairs only, and no Memora ground-truth
label enters the objective. Exact group, row, and source-text overlap between the
synthetic training pool and the Memora evaluation set is zero. Two couplings remain
at the level of surface form rather than supervision, and we state them directly.
First, five entity names (a small set of public-figure and topic strings) appear
in both pools as surface strings; the Memora evaluation instances and their labels
stay outside training. Second, the synthetic hard-negative templates
(\S\ref{sec:strat}) are designed from an error analysis of Memora role-binding
failures, which transfers the \emph{shape} of the hard cases without transferring
their labels. The transfer is therefore label-free, and the role-binding numbers
hold under this exact sense.

\subsection{The deployed query-conditioned calibrator}
\label{sec:calibrator}

The released \texttt{demote} mode does not deploy the synthetic-graph scorer. It
uses a query/source/target cross-encoder calibrator whose input is conditioned on
the user query:
\begin{quote}\small
\texttt{USER\_QUERY: \{query\}} \\
\texttt{SOURCE\_EVIDENCE: \{source\}} \\
\texttt{TASK: Decide whether the target memory should be demoted for this user
query.}
\end{quote}
paired with \texttt{TARGET\_MEMORY: \{target\}}. It is trained with a weighted
cross-entropy plus an event-role margin term,
\begin{equation}
\mathcal{L} \;=\; \mathrm{CE}_{w}\big(\hat{y}, y\big)
\;+\; \lambda_{\mathrm{role}} \cdot \mathcal{L}_{\mathrm{role\_margin}},
\label{eq:loss-cal}
\end{equation}
where $\mathcal{L}_{\mathrm{role\_margin}}$ requires that, within one event group,
the rows that should be demoted (old or current bindings under update) score above
the rows that must be protected. The margin term is what makes the calibrator
query-conditioned rather than query-blind, which is the property that the
never-demote and query-blind baselines lack (\S\ref{sec:dense}).

\section{Experimental Protocol}
\label{sec:protocol}

\paragraph{Settings.} Real-distribution validity behavior is evaluated on Memora
\cite{memora2026}, a public long-term memory benchmark introduced by Uddin et al.,
spanning weeks-to-months user conversations with explicit memory mutation and a
Forgetting-Aware Memory Accuracy (FAMA) metric that penalizes reliance on obsolete
memory. The A1 role-binding and dense-retrieval demotion slices used here are
derived from the released Memora traces. Memora labels are not used as training
targets; Memora residual errors are used only to identify failure patterns for
synthetic template design (\S\ref{sec:feedback}).
Three evaluation settings are used. The \emph{synthetic
multi-hop validity benchmark} constructs invalidation chains of controlled length,
where an update at one node should invalidate downstream nodes that depend on it;
it is the setting in which the dual-evidence gate and the propagation mechanism
are measured with known structure. The \emph{Memora A1 role-binding transfer}
setting evaluates whether a candidate is the superseded binding of a target whose
attribute a later source rebinds; it is the real-data setting and uses zero
target-side training labels. The \emph{Memora dense current-state retrieval}
setting asks for a target's current attribute against a store containing both the
current binding and superseded ones; it is the setting in which the deployed
validity layer is measured end-to-end.

\paragraph{Seeds and aggregation.} Method-level tables use seeds
$[7, 11, 23, 31, 47]$ and report mean $\pm$ standard deviation. The dense-retrieval
method-level evaluation scores $69{,}200$ source-query rows across the five seeds,
with $34{,}600$ top-1 summary rows. The released-checkpoint benchmark is a single
checkpoint over $6{,}920$ source-query rows and $20{,}760$ target predictions: each
source-query row is scored against the candidate targets in its group, so one row
expands to several \texttt{(query, source, target)} scoring instances (here
$20{,}760 / 6{,}920 = 3$ on average). The
Memora A1 role-binding set has 173 groups, with a 37-group rule-hard subset and a
12-group both-failed subset; the wider intervals on these subsets reflect their
size.

\paragraph{Metrics.} On role binding the metric is group-all-correct accuracy: a
group counts as correct only when every binding decision in the group is correct.
On dense retrieval the metrics are pair accuracy (target-conditioned relation
accuracy), demote recall (fraction of superseded memories correctly demoted),
protect recall (fraction of non-superseded memories correctly preserved), and
current-active Hit@1 (H@1, end-task retrieval of the current binding at rank 1). On the
synthetic benchmark the metric is overall accuracy, reported jointly with
invalidated-case and valid-case accuracy so that aggressive propagation cannot
trade valid-case correctness for invalidated-case recall undetected.

\paragraph{Threshold selection.} The dual-evidence gate selects a decision
threshold per seed and per fold on a validation split, optimizing overall accuracy
jointly with invalidated-case and valid-case accuracy
(\S\ref{sec:noisyor}). Thresholds are never selected on the test split.

\paragraph{Isolation.} Evaluation-only fields (gold labels, staleness flags,
teacher scores, GPT labels, and entity/slot identifiers) are used only as training
or evaluation targets and never as inference features; the deployed API rejects
them at inference (\S\ref{sec:contracts}). The synthetic-graph scorer and the
deployed query-conditioned calibrator are trained separately, and the calibrator
uses no synthetic-graph labels at inference.

\section{Verifier Progression}
\label{sec:progression}

\begin{figure}[htbp]
\centering
\includegraphics[width=0.95\textwidth]{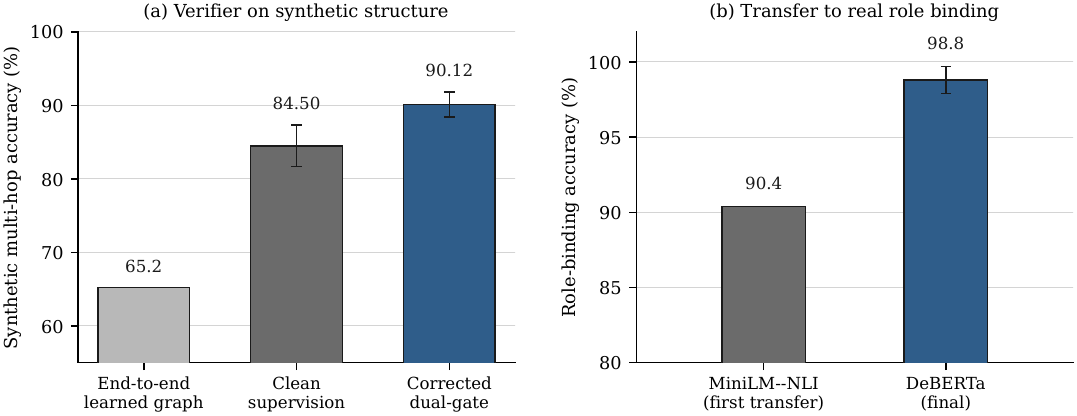}
\caption{Verifier progression. (a) Synthetic multi-hop accuracy across milestones:
the $65.2\%$ end-to-end learned-graph baseline, the $84.5\% \pm 2.8$
clean-supervision milestone, and the $90.12\% \pm 1.73$ corrected dual-evidence
gate. (b) Memora role-binding transfer (full set, zero target-side labels) from
the first transfer checkpoint ($90.4\%$) to the final DeBERTa verifier
($98.8\% \pm 0.9$), showing how the synthetic-side mechanism converts into
real-data transfer.}
\label{fig:milestones}
\end{figure}

Figure~\ref{fig:milestones} traces the verifier across three milestones on the
synthetic benchmark. The first end-to-end system that turned text into a
dependency graph and propagated over it reached an overall $65.2\%$, with visible
edge-error amplification along chains. Clean real-slot supervision
(\S\ref{sec:strat}) lifted accuracy to $84.5\% \pm 2.8$. The corrected
dual-evidence gate, which fixed a duplicate-edge deduplication bug and applies
Equations~\eqref{eq:slot}--\eqref{eq:noisyor}, reached $90.12\% \pm 1.73$, the
synthetic headline. A nearby run measured $90.20\% \pm 2.19$ before this
correction; the corrected $90.12\% \pm 1.73$ is the figure used throughout.

\section{Role-Binding Verification}
\label{sec:rolebinding}

The strongest validated capability is target-conditioned role binding: deciding,
for a target whose attribute a later source rebinds, whether a candidate memory is
the superseded binding. Table~\ref{tab:rolebind} reports group-all-correct
accuracy on the Memora A1 transfer set (173 groups), the rule-hard subset (37
groups where a target-position rule fails by construction), and the both-failed
subset (12 groups where the position rule and zero-shot NLI both fail). All
trained rows use zero Memora target-side labels.

\begin{table}[htbp]
\centering
\caption{Target-conditioned role binding on Memora A1 transfer (group-all-correct
accuracy). all-173 is the full set; rule-hard-37 is the subset where a
target-position rule fails by construction; both-failed-12 is the subset where the
position rule and zero-shot NLI both fail. v3 rows use zero Memora target-side
training labels; final rows are 5-seed mean $\pm$ std. A dash (---) marks a split
that is undefined for that baseline.}
\label{tab:rolebind}
{\setlength{\tabcolsep}{5pt}
\begin{tabular}{lrrr}
\toprule
Method & all-173 & rule-hard-37 & both-failed-12 \\
\midrule
Target-position rule              & 78.6\%  & 0.0\%   & --- \\
Zero-shot NLI                     & 64.2\%  & 67.6\%  & 0.0\% \\
MS MARCO MiniLM CE (relevance)    & 17.9\%  & ---     & --- \\
MiniLM--NLI (first transfer)      & 90.4\%  & 81.1\%  & 66.7\% \\
DeBERTa $+$ merged data (final)   & \textbf{98.8\% \plusminus 0.9} & \textbf{96.2\% \plusminus 4.1} & \textbf{93.3\% \plusminus 9.1} \\
\bottomrule
\end{tabular}}
\end{table}

The verifier holds up exactly where the cheap signals collapse. A target-position
rule scores $78.6\%$ overall but $0\%$ on the rule-hard subset, which is hard by
construction; zero-shot NLI scores $64.2\%$ overall and $0\%$ on the both-failed
subset. The relevance cross-encoder scores $17.9\%$, confirming that relevance
ranking and validity verification are distinct tasks. The first transfer
checkpoint already reached $90.4 / 81.1 / 66.7$ across the three splits; the final
DeBERTa verifier with merged supervision reaches $98.8 / 96.2 / 93.3$, with the
wider intervals on the small hard subsets reflecting their size (37 and 12
groups).

\section{Multi-Hop Validity Propagation}
\label{sec:multihop}

\begin{figure}[htbp]
\centering
\includegraphics[width=0.78\textwidth]{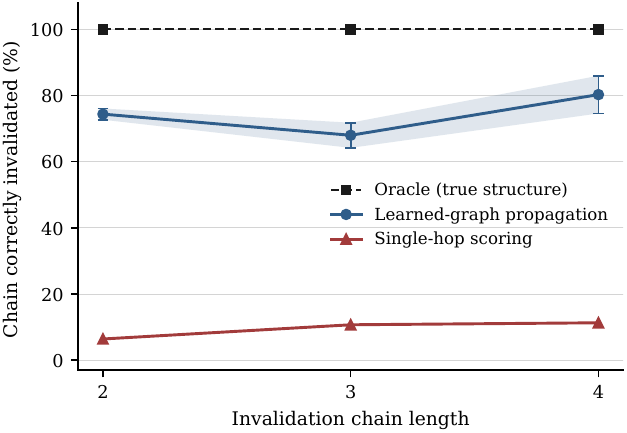}
\caption{Multi-hop validity propagation on synthetic invalidated chains
(invalidated-only). Single-hop scoring captures a small, roughly flat fraction of
the chain; propagation over a learned graph recovers most of it; oracle
propagation over the true structure is the ceiling.}
\label{fig:multihop}
\end{figure}

When an update invalidates a chain of dependent memories, validity must propagate
along the dependency structure. Figure~\ref{fig:multihop} and
Table~\ref{tab:multihop} report a diagnostic on synthetic invalidated chains of
length 2, 3, and 4: the fraction of the chain correctly marked invalid by
single-hop scoring, by propagation over a learned graph, and by an oracle that
propagates over the true structure.

\begin{table}[htbp]
\centering
\caption{Multi-hop validity propagation. The upper block reports all chains by
length (recency, v2-relevance, single-text, single-hop, learned-graph, oracle);
the lower block reports the invalidated-only fraction correctly invalidated.
Learned-graph rows are 5-seed mean $\pm$ std. Per-chain sample sizes are: all
chains $n = 159 / 170 / 173$ at lengths $2/3/4$; invalidated-only
$n = 110 / 75 / 62$; valid controls $n = 49 / 95 / 111$.}
\label{tab:multihop}
{\setlength{\tabcolsep}{5pt}
\begin{tabular}{lrrr}
\toprule
Method & chain-2 & chain-3 & chain-4 \\
\midrule
\multicolumn{4}{l}{\emph{All chains}} \\
recency               & 38.4 & 45.3 & 56.1 \\
v2 relevance          & 29.6 & 36.5 & 42.8 \\
single-text           & 59.1 & 50.0 & 53.2 \\
single-hop            & 35.2 & 60.6 & 68.2 \\
learned-graph         & \textbf{70.2 \plusminus 1.8} & \textbf{64.5 \plusminus 4.7} & \textbf{61.3 \plusminus 4.6} \\
oracle                & 100  & 100  & 100 \\
\midrule
\multicolumn{4}{l}{\emph{Invalidated-only}} \\
single-hop            & 6.4  & 10.7 & 11.3 \\
learned-graph         & \textbf{74.4 \plusminus 1.7} & \textbf{68.0 \plusminus 3.8} & \textbf{80.3 \plusminus 5.7} \\
oracle                & 100  & 100  & 100 \\
\bottomrule
\end{tabular}}
\end{table}

On the invalidated-only split, single-hop scoring captures only a small fraction
of the chain (6--11\%) regardless of length, because a single-hop judgment cannot
see that an upstream update has invalidated a node two or three hops away.
Propagation over a learned graph recovers most of the chain (68--80\%),
demonstrating that propagation is the right structure for multi-hop validity. The
two blocks trend in opposite directions with chain length, and the reason is the
shifting composition of each split: as chains lengthen, the all-chains set is
increasingly dominated by valid controls (the valid-control count rises from 49 to
111 across lengths $2$ to $4$), which pulls the all-chains learned-graph figure
down, while the invalidated-only set shrinks to $n = 62$ at length 4. The
invalidated-only rebound to $80.3\%$ at length 4 should not be over-read: it rests
on the smallest per-chain sample and carries the widest interval ($\pm 5.7$),
consistent with small-sample variation rather than a genuine increase in
propagation accuracy with depth. The gap to the oracle, together with
false-overturn harm on the valid controls, scopes propagation to the
given-structure regime: with a reliable dependency structure, propagation works;
constructing that structure fully automatically is the boundary characterized in
\S\ref{sec:findings}.

\section{The Validity Context Layer in Dense Retrieval}
\label{sec:dense}

The deployed layer is evaluated on Memora dense current-state retrieval, where a
query asks for a target's current attribute and the store contains both the
current binding and superseded ones. Table~\ref{tab:dense} reports method-level
results (5 seeds, $69{,}200$ source-query rows scored across seeds), with the
released-checkpoint figures reported alongside in \S\ref{sec:released}.

\begin{table}[htbp]
\centering
\caption{Validity context layer on Memora dense current-state retrieval (5-seed
mean $\pm$ std). never-demote is the no-validity baseline; query-blind is a
relation classifier that ignores the target and matches never-demote exactly;
top1-retrieved-source is the deployed policy; oracle-source uses the true
superseding source. pair acc is target-conditioned relation accuracy;
demote/protect recall measure superseded vs non-superseded handling;
current-active H@1 is end-task retrieval of the current binding.}
\label{tab:dense}
{\setlength{\tabcolsep}{4pt}
\begin{tabular}{lrrrr}
\toprule
Method & pair acc & demote recall & protect recall & current-active H@1 \\
\midrule
never-demote                & 87.5\% & 0.0\%   & 100\%   & 45.1\% \\
query-blind relation        & 87.5\% & 0.0\%   & 100\%   & 45.1\% \\
top1-retrieved-source (v3)  & \textbf{98.6\% \plusminus 0.2} & \textbf{92.9\% \plusminus 1.1} & \textbf{99.4\% \plusminus 0.1} & \textbf{95.7\% \plusminus 1.2} \\
oracle-source               & 99.9\% & ---     & ---     & 100\% \\
\bottomrule
\end{tabular}}
\end{table}

The never-demote baseline retrieves the current binding only $45.1\%$ of the time,
because superseded memories outrank the current one on relevance alone. A
query-blind relation classifier, one that judges the relation without
conditioning on the target, matches the never-demote numbers exactly: knowing
the relation without the query intent collapses to protecting every memory ($0\%$
demote recall, $100\%$ protect recall), so it never recovers the current binding.
This is the empirical case for target conditioning: the relation signal is only
useful once it is tied to the target the query asks about. The deployed v3 layer,
using the top-1 retrieved source as the superseding evidence, raises
current-active H@1 to $95.7\% \pm 1.2$ while protecting non-superseded memories at
$99.4\%$ recall and demoting superseded ones at $92.9\%$ recall. The oracle source
(using the true superseding memory) reaches $100\%$ H@1, locating the remaining
headroom in source selection rather than in the relation judgment itself.

\paragraph{Old-target and event splits.} On the same setting the layer reaches
$92.8\% \pm 1.1$ on the old-target all-type split and $89.1\% \pm 1.3$ on the
event all-type split, consistent with role binding being the strongest regime and
event-style updates the harder one.

\section{LoCoMo Density Audit}
\label{sec:locomo}

ConvMemory v1 and v2 are evaluated on LoCoMo \cite{locomo2024}, so a natural question is how the
validity layer behaves there. The answer is a finding about the data rather than a
benchmark result: on LoCoMo, supersession is rare enough that the layer's
correct default is to annotate, not to demote. A high-value GPT audit of 100
candidate cases found valid demotions in $3$ of them ($3.0\%$ yield), and $20$
negative controls produced $0$ false positives. The audited positive set is sparse
(on the order of a handful of confidently superseded memories across the audit),
which is too thin to support a method-level demotion benchmark. The audit is used
to decide deployment scope, not to estimate method-level accuracy. Accordingly, the
\texttt{context} default preserves ranking on LoCoMo with no behavior change, and
the opt-in \texttt{demote} mode shows no reliable gain there. LoCoMo is therefore
evidence that real conversational supersession density can be low (exactly the
regime in which annotate-by-default is the safe policy), rather than
a validation set for demotion. The settings where demotion is validated (Memora
dense current-state retrieval) are those with enough supersession density to
measure it.

\section{Cost-Conscious Routing}
\label{sec:b1}

This section analyzes the research dual-gate verifier, not the released
single-checkpoint calibrator of \S\ref{sec:released}. The dual-evidence gate runs
both slot heads on every pair. A cost-conscious
router (B1) instead runs the cheap MiniLM head first and routes to the DeBERTa
head only when the cheap head is uncertain. On the synthetic benchmark, the best
router (a fast-accept variant with agreement gating, combined by noisy-or) reaches
$89.56\% \pm 2.54$, indistinguishable within seed-level variance from the
full-product anchor in the same run ($89.40\% \pm 2.79$) and from the corrected
gate ($90.12\% \pm 1.73$).

The cost behavior, however, depends on the data, not the router. On the synthetic
benchmark the router still sends $99.5\%$ of pairs to the DeBERTa head, because
synthetic data has almost no easy mass by construction. On real Memora data the
router preserves the transfer numbers ($90.4 / 81.1 / 66.7$) while sending far
fewer pairs to the expensive head: $75.0\%$ at the tightest agreement threshold,
$86.0\%$ at a looser one, and $94.5\%$ at the loosest. The reframing follows
directly: \emph{cost saving depends on the easy mass in the deployment
distribution, not on the router alone}. A router added to a distribution with no
easy mass saves nothing; the same router on real data, which contains easy mass,
routes a meaningful fraction away from the expensive head. Quantified against the
component latencies of Table~\ref{tab:latency-config}, routing on real data
reduces the full dual-gate cost from about $18.9$ to about $15.1$ ms per top-10
query, while on synthetic data the near-total routing leaves it essentially
unchanged at about $18.8$ ms.

\section{Deployment: Modes, Contracts, and Cost}
\label{sec:deploy}

\subsection{Modes}
\label{sec:modes}

The layer exposes four accepted values through a single \texttt{validity\_mode}
argument, corresponding to three behavioral modes (\texttt{None} and \texttt{off}
share one behavior).

\begin{itemize}
    \item \texttt{None} / \texttt{off}: ConvMemory behavior is unchanged; the
    v1/v2 path is byte-identical to a build without the v3 module.
    \item \texttt{context} (default): the layer attaches structured validity
    metadata to each result while preserving the candidate set and the rank order.
    The agent reads the validity context note and the
    \texttt{possibly\_outdated} flag and decides how to use them.
    \item \texttt{demote}: an explicit opt-in that reorders results for dense
    current-state workloads, preserving the candidate set and result count while
    moving superseded memories down. This is the only mode in which v3 changes
    ordering.
\end{itemize}

\subsection{Safety contracts}
\label{sec:contracts}

Six properties are enforced by named tests rather than asserted in prose. The full
suite reaches 41 passing tests. Table~\ref{tab:contracts} maps each contract to
its enforcing test.

\begin{table}[htbp]
\centering
\caption{Validity-layer safety contracts, each enforced by a named test. The full
suite reaches 41 passing tests; the supplementary tests below the rule cover
batching, format, and round-trip behavior.}
\label{tab:contracts}
\renewcommand{\arraystretch}{1.18}
\footnotesize
\begin{tabular}{>{\raggedright\arraybackslash}p{0.40\textwidth} >{\raggedright\arraybackslash}p{0.52\textwidth}}
\toprule
Contract & Enforcing test \\
\midrule
\texttt{off}/\texttt{None} unchanged & \path{test_off_mode_byte_identical} \\
\texttt{context} preserves rank order & \path{test_context_mode_preserves_ranking} \\
\texttt{demote} preserves candidate set & \path{test_demote_preserves_candidate_set} \\
\texttt{demote} is explicit opt-in & \path{test_demote_is_opt_in} \\
Forbidden inputs rejected & \path{test_forbidden_fields_rejected} \\
Context evidence sanitized & \path{test_context_evidence_has_no_forbidden_fields} \\
\midrule
Save/load round-trip & \path{test_validity_save_load_roundtrip} \\
Query/source/target format & \path{test_validity_cross_encoder_uses_query_source_target_format} \\
Batched apply-pairs & \path{test_validity_cross_encoder_batches_apply_pairs} \\
Explicit-pair batching & \path{test_validity_score_evidence_pairs_batches_explicit_pairs} \\
Cross-encoder round-trip & \path{test_validity_cross_encoder_save_load_roundtrip} \\
Invalid mode raises & \path{test_invalid_validity_mode_raises} \\
\bottomrule
\end{tabular}
\end{table}

The input contract rejects a fixed set of evaluation-only fields. If any candidate
passed at inference contains any of the following keys, the API raises
\texttt{ValueError}:
\begin{quote}\small
\texttt{gold}, \texttt{gold\_ids}, \texttt{is\_current}, \texttt{is\_latest},
\texttt{is\_stale}, \texttt{stale}, \texttt{answer}, \texttt{answer\_text},
\texttt{ce\_score}, \texttt{mxbai\_score}, \texttt{teacher\_score},
\texttt{gpt\_label}, \texttt{entity\_id}, \texttt{slot\_id}.
\end{quote}
The output contract sanitizes context evidence so that none of these fields leaks
into the validity metadata returned to the agent.

\subsection{Cost and configuration}
\label{sec:cost}

The released default checkpoint is a DeBERTa-v3 NLI backbone with
$184{,}423{,}682$ parameters. The default package mode is
\texttt{validity\_mode="context"}; demotion is an explicit opt-in. The source
policy is top-1. As a released package measurement, the batched public API at
batch size 512 scores $6{,}920$ source-query rows at $1.5844$ ms per source-query
pair, with a module load time of $2.16$ s. An earlier unbatched smoke path through
the same module measured $17.587$ ms per pair and is retained only as a
non-headline sanity check; the batched figure is the headline cost. Annotating a
top-10 context costs on the order of $10 \times 1.5844$ ms of scorer time before
retrieval and source-selection overhead; the measured unit is per pair.

\begin{table}[htbp]
\centering
\caption{Released package configuration and measured cost (public API, batch size
512). The public API unit is one source-query row; internal scoring expands to
\texttt{(query, source, target)} instances.}
\label{tab:cost}
\renewcommand{\arraystretch}{1.15}
\begin{tabular}{>{\raggedright\arraybackslash}p{0.50\textwidth} >{\raggedright\arraybackslash}p{0.42\textwidth}}
\toprule
Item & Value \\
\midrule
Backbone & DeBERTa-v3 NLI, $184{,}423{,}682$ params \\
Default mode & \texttt{validity\_mode="context"} \\
Demotion & explicit opt-in \\
Source policy & top-1 \\
API batch size & 512 \\
Pairs per target query & $1.0$ \\
Target predictions per source-query row & $3.0$ avg \\
Scoring cost & $1.5844$ ms / source-query pair \\
Module load time & $2.16$ s \\
\bottomrule
\end{tabular}
\end{table}

\paragraph{Component latency and configuration estimates.} A separate component
latency probe measures each slot head in isolation across batch sizes, which
supports estimating the dual-gate and routed-gate costs from measured components.
This probe is distinct from the released package measurement above: it scores a
fixed probe set per batch size rather than running the full public API benchmark,
so its per-pair figures are component timings rather than a single deployed cost.
Table~\ref{tab:latency-comp} reports the component timings, and
Table~\ref{tab:latency-config} reports the configuration estimates built from
them.

\begin{table}[htbp]
\centering
\caption{Component scoring latency from the component probe. Each slot head is
timed in isolation; ``best batched'' is the fastest batch size found for that
head in this probe.}
\label{tab:latency-comp}
\renewcommand{\arraystretch}{1.15}
\begin{tabular}{lrrr}
\toprule
Scorer & batch$=1$ & best batched & speedup \\
\midrule
MiniLM-only      & $4.144$ ms/pair  & $0.253$ ms/pair (@128) & about $16.4\times$ \\
DeBERTa released & $16.498$ ms/pair & $1.499$ ms/pair (@32)  & about $11.0\times$ \\
\bottomrule
\end{tabular}
\end{table}

\begin{table}[htbp]
\centering
\caption{Configuration cost estimated from the component latencies of
Table~\ref{tab:latency-comp} (batch$=32$). The dual-gate and routed-gate rows are
estimated as $\mathrm{dual}=\mathrm{MiniLM}+\mathrm{DeBERTa}$ and
$\mathrm{routed}=\mathrm{MiniLM}+r\cdot\mathrm{DeBERTa}$ for routing fraction $r$,
not measured as direct end-to-end wall-clock.}
\label{tab:latency-config}
\renewcommand{\arraystretch}{1.15}
\begin{tabular}{lrr}
\toprule
Configuration & ms/pair (batch$=32$) & top-10/query \\
\midrule
MiniLM-only                       & $0.389$ & $3.89$ \\
DeBERTa-only                      & $1.499$ & $14.99$ \\
full dual (MiniLM $+$ DeBERTa)    & $1.888$ & $18.88$ \\
routed, real data ($75\%$ route)  & $1.513$ & $15.13$ \\
routed, synthetic ($99.5\%$ route)& $1.880$ & $18.80$ \\
\bottomrule
\end{tabular}
\end{table}

The dual-gate and routed-gate figures are estimated from the measured MiniLM and
DeBERTa component latencies rather than measured as a single end-to-end wall-clock
run. On real data, where a quarter of pairs are routed away from the expensive
head, the routed gate is estimated at $15.13$ ms per top-10 query against the full
dual gate's $18.88$; on synthetic data, where nearly all pairs reach the expensive
head, the routed gate stays at $18.80$, essentially unchanged. This quantifies the
easy-mass effect of \S\ref{sec:b1}: the routing saving tracks the easy mass of the
deployment distribution.

\subsection{Released checkpoint figures}
\label{sec:released}

The released checkpoint reproduces the method-level dense-retrieval behavior. On
the public API/package benchmark (single checkpoint, $6{,}920$ source-query rows,
$20{,}760$ target predictions) it reaches pair accuracy $98.7\%$, demote recall
$93.6\%$, protect recall $99.4\%$, old-target all-type $93.1\%$, event all-type
$89.6\%$, and current-active H@1 $96.5\%$, at $1.5844$ ms per source-query pair.
These are the figures a user reproduces from the published checkpoint; the figures
in Table~\ref{tab:dense} are the 5-seed method-level estimates. The provenance
relationship is detailed in Appendix~\ref{app:provenance}.

\section{Boundary Findings}
\label{sec:findings}

Two boundaries are reported as findings, each scoping a capability precisely.

\paragraph{Multi-hop relation labels for free-form conversation are scarce.} The
verifier is validated on role binding with real transfer (Memora) and on multi-hop
propagation with synthetic structure. Target-conditioned relation labels for
free-form multi-hop conversation (the data needed to validate end-to-end
multi-hop propagation on natural dialogue) are not yet available at the scale
required, to our knowledge. v3 therefore validates multi-hop propagation in the
given-structure regime and reports natural-dialogue multi-hop validation as the
next data milestone.

\paragraph{Strict prerequisite edges require counterfactual necessity.} An
extensive line of experiments on automatic dependency-graph construction concluded
that strict prerequisite edges (edges asserting that one memory is necessary for
another) require a counterfactual necessity judgment grounded in world
knowledge. A discriminative relation model provides support edges reliably but does
not provide strict necessity reliably, and aggressive automatic propagation
introduces false-overturn harm on valid controls (\S\ref{sec:multihop}). Automatic
strict-prerequisite graph construction is therefore not part of the released
default path; it is reported as a boundary and future research direction, and the
deployed layer uses given or support structure rather than inferred strict
prerequisites.

\paragraph{Demotion is scoped, not default.} Query-conditioned demotion is
validated on Memora dense current-state retrieval, where a query asks for a
target's current attribute. On general conversational retrieval the safe default is
\texttt{context}: surface the validity evidence and let the agent decide. Demotion
is an explicit opt-in for the dense current-state regime where it is validated.

\section{Reproducibility}
\label{sec:repro}

The package is installable via \texttt{pip install convmemory==0.6.0} (PyPI). The
released validity checkpoint is on the Hugging Face Hub at
\path{Purdy0228/ConvMemory-v3-Validity-Context}. The build produces a clean
46~KiB wheel\footnote{The wheel is $47{,}504$ bytes.} with no results, checkpoints,
or weights mixed in, under the MIT license. A minimal usage example:

\begin{verbatim}
from convmemory import ConvMemory
model = ConvMemory.from_pretrained("Purdy0228/ConvMemory-LoCoMo-MPNet")
model.load_validity_module("Purdy0228/ConvMemory-v3-Validity-Context")
# context mode (default): preserves ranking, attaches validity metadata
results = model.retrieve(
    query=q,
    memories=ms,
    top_k=10,
    validity_mode="context",
)
for r in results:
    if r.validity and r.validity.get("status") == "possibly_outdated":
        print(r.validity["context_note"]) 
# validity_mode="demote" is an explicit opt-in that reorders results
\end{verbatim}

\section{Discussion}
\label{sec:discussion}

ConvMemory v3 contributes a validity context layer: target-conditioned relation
verification, conservative evidence surfacing, and opt-in query-conditioned
demotion. The verifier transfers from synthetic structure to real role binding
with zero target-side labels, the deployed layer preserves retrieval by default
and recovers current-state accuracy under explicit opt-in, and its behavior is
pinned by machine-verifiable contracts. Multi-hop graph propagation is validated as
a mechanism; full automatic strict dependency-graph construction remains a
boundary, because strict necessity requires counterfactual world knowledge.

\paragraph{Why target conditioning is the load-bearing idea.} Three independent
results point at the same conclusion. The architecture ablation
(\S\ref{sec:merge}) shows that removing the slot interaction costs about 14 points
while merging the event and operation heads costs nothing, so the
target-conditioned slot product, not the event/operation machinery, carries the
signal. The dense-retrieval comparison (\S\ref{sec:dense}) shows that a query-blind
relation classifier collapses to the never-demote failure shape, so the relation
signal is only useful once tied to the target. The role-binding comparison
(\S\ref{sec:rolebinding}) shows that a relevance cross-encoder scores $17.9\%$, so
relevance and validity are genuinely different tasks. Across the three settings,
the property that separates a working verifier from a failing one is the same:
conditioning the relation judgment on the specific target proposition.

\paragraph{Where the headroom is.} In dense retrieval the oracle source reaches
$100\%$ current-active H@1 against the deployed layer's $95.7\%$, which places the
remaining headroom in source selection rather than in the relation judgment. The
relation judgment itself is near-saturated (pair accuracy $98.6\%$); the path to
the oracle runs through retrieving the correct superseding source, not through a
stronger relation model.

\subsection{Limitations}
\label{sec:limitations}

\paragraph{Multi-hop validation uses synthetic structure.} The propagation
mechanism is validated on synthetic invalidation chains with known structure.
End-to-end validation on natural multi-hop dialogue awaits target-conditioned
relation labels at scale, which are, to our knowledge, scarce (\S\ref{sec:findings}).

\paragraph{Strict-edge construction is a boundary, not a released feature.} The deployed layer uses given
or support structure. Fully automatic construction of strict prerequisite edges is
not part of the released default path and is reported as a boundary and future
research direction, because strict necessity requires counterfactual world
knowledge that a discriminative model does not provide reliably.

\paragraph{Demotion is scoped to dense current-state retrieval.} The
\texttt{demote} mode is validated on Memora dense current-state retrieval and is an
explicit opt-in there. On general conversational retrieval the validated default is
\texttt{context}, which surfaces evidence without reordering.

\paragraph{Cost savings are distribution-dependent.} The cost-conscious router
(\S\ref{sec:b1}) saves compute in proportion to the easy mass of the deployment
distribution. On distributions with little easy mass, both heads run on most pairs
and the saving is small; the saving is a property of the data, reported as such.

\paragraph{Released checkpoint versus method-level figures.} The published
checkpoint is a single representative checkpoint; the headline dense-retrieval
figures are 5-seed method-level estimates (Appendix~\ref{app:provenance}). A user
evaluating the single checkpoint reproduces the package-benchmark figures, which
are consistent with, and measured separately from, the 5-seed estimates.

\subsection{Future Work}
\label{sec:future}

Three directions follow. First, target-conditioned relation labels on natural
multi-hop dialogue would lift multi-hop propagation from the given-structure regime
to end-to-end validation. Second, source selection is the located headroom in dense
retrieval: a stronger source-retrieval policy converts directly into end-task
accuracy, since the relation judgment is already near-saturated. Third, the
validity layer composes with an agent's memory-management loop, where the
\texttt{context} annotations become inputs to write, read, and consolidate
decisions.

\appendix

\section{Source-of-Truth Checklist}
\label{app:sources}

Every quantitative result in this report traces to an internal measurement record.
Table~\ref{tab:sources} maps result types to the measurement that establishes them.
The released package and checkpoint are public (PyPI and the Hugging Face Hub);
the underlying measurement records are local artifacts retained by the author and
available on request.

\begin{table}[H]
\centering
\caption{Mapping from result type to the measurement that establishes it.}
\label{tab:sources}
\renewcommand{\arraystretch}{1.2}
\begin{tabular}{>{\raggedright\arraybackslash}p{0.46\textwidth} >{\raggedright\arraybackslash}p{0.46\textwidth}}
\toprule
Result & Source measurement \\
\midrule
End-to-end learned-graph baseline ($65.2\%$) & initial text-to-graph propagation run \\
Slot input form / MiniLM head & alignment-form balanced slot training \\
DeBERTa NLI-backbone slot head & NLI-backbone slot probe \\
Corrected dual-evidence gate ($90.12\%$) & corrected dual-verifier-gate run \\
Old three-factor edge formula & fixed-formula edge ablation \\
Event/operation head merge ($+0.00\%$, CI) & event/operation merge significance test \\
Role-binding baselines (rule / NLI / CE) & zero-shot baseline and first-transfer runs \\
First zero-label transfer ($90.4/81.1/66.7$) & first role-binding transfer checkpoint \\
Final Memora role binding ($98.8/96.2/93.3$) & final DeBERTa merged-data evaluation \\
Dense demotion, method-level (5-seed) & dense current-state method-level evaluation \\
Query-conditioned calibrator objective & event-level role-binding calibrator training \\
Released checkpoint / package benchmark & released checkpoint and public API benchmark \\
Component / configuration latency & component latency probe \\
\bottomrule
\end{tabular}
\end{table}

\clearpage
\section{Released Checkpoint Provenance}
\label{app:provenance}

The relationship between the published checkpoint and the headline numbers follows
the same three-level convention as ConvMemory v2.

\begin{itemize}
    \item \textbf{Method-level}: 5-seed estimates over seeds
    $[7, 11, 23, 31, 47]$, $69{,}200$ source-query rows scored across seeds, with
    top-1 summary rows numbering $34{,}600$. These are the headline dense-retrieval
    figures (Table~\ref{tab:dense}).
    \item \textbf{Representative checkpoint}: the seed-7 representative checkpoint
    exported for release.
    \item \textbf{Package benchmark}: the public API/package benchmark for that
    checkpoint, $6{,}920$ source-query rows and $20{,}760$ target predictions
    (\S\ref{sec:released}).
\end{itemize}

The released checkpoint implements the method; the method-level figures are 5-seed
estimates, and the package-benchmark figures are the single-checkpoint
measurements a user reproduces. The recorded artifact hashes are
\begin{quote}\small
model: \texttt{446ee0cf6df4a8967e1a78c46d2ff3a2d777de65efbf475d2278d99468faa8d9} \\
config: \texttt{81eddb5f2ff4545dcf4b7655fedd1f7cf846248ad8962394195e6960a2e07849}
\end{quote}

\end{document}